\documentclass{article}
\usepackage[final,nonatbib]{proposal}
\usepackage[utf8]{inputenc} 
\usepackage[T1]{fontenc}    
\usepackage{hyperref}       
\usepackage{url}            
\usepackage{booktabs}       
\usepackage{amsfonts}       
\usepackage{nicefrac}       
\usepackage{microtype}      
\usepackage{xcolor}         

\usepackage{graphicx}
\usepackage{verbatim}
\usepackage{multirow}
\usepackage{witharrows}
\usepackage{color}
\usepackage{amsmath}
\usepackage{mathtools}

\usepackage{times}
\usepackage{epsfig}
\usepackage{graphicx}
\usepackage{amsmath}
\usepackage{amssymb}
\usepackage{float}

\usepackage{amsfonts}
\usepackage{enumerate}
\usepackage{amsmath}
\usepackage{booktabs}
\usepackage{color}
\usepackage{array}
\usepackage{multirow}
\usepackage{multicol}
\usepackage{graphicx}  
\usepackage{url}
\usepackage{bbding}
\usepackage{cuted}
\usepackage{capt-of}
\usepackage[font=footnotesize,labelfont=bf]{caption}
\usepackage{footnote}
\usepackage{caption}
\usepackage{subfigure}

\title{18-786 Project Final Report \\ Object  Segmentation with Audio Context}

\author{%
 Kaihui Zheng\\
 Electrical and Computer Engineering\\
 Carnegie Mellon University\\
 \texttt{kaihuiz@andrew.cmu.edu} \\
  \And
  Yuqing Ren \\
  Electrical and Computer Engineering\\
  Carnegie Mellon University\\
  \texttt{yuqingr@andrew.cmu.edu} \\
  \AND
  Zixin Shen \\
  Electrical and Computer Engineering\\
  Carnegie Mellon University\\
  \texttt{zixins@andrew.cmu.edu} \\
  \And
  Tianxu Qin \\
  Electrical and Computer Engineering\\
  Carnegie Mellon University\\
  \texttt{tianxuq@andrew.cmu.edu} \\
}

\begin{document}
\maketitle
\begin{abstract}
Visual objects often have acoustic signatures that are naturally synchronized with them in audio-bearing video recordings. For this project, we explore the multimodal feature aggregation for video instance segmentation task, in which we integrate audio features into our video segmentation model to conduct an audio-visual learning scheme. Our method is based on existing video instance segmentation method which leverages rich contextual information across video frames. Since this is the first attempt to investigate the audio-visual instance segmentation, a novel dataset, including 20 vocal classes with synchronized video and audio recordings, is collected. By utilizing combined decoder to fuse both video and audio features, our model shows a slight improvements compared to the base model. Additionally, we managed to show the effectiveness of different modules by conducting extensive ablations.
\end{abstract}

\section{Introduction}
Benefiting from the development in the video domain, video instance segmentation (VIS) emerges and attracts lots of attention. VIS aims to segment object masks on individual frames while keeping the identity consistent throughout the entire video. The previous methods focusing on improving segmentation quality from sole video modality while ignoring the strong information contained in the corresponding audio recording. In this report, not only will we investigate the VIS task by considering both visual and auditory signals simultaneously, but also we are going to implement our model with different structures.

For single-modal audio-based perception problem, previous methods have achieved promising performance on several tasks, such as event detection \cite{1}, speech recognition \cite{2} and sound classification \cite{3}. For single-modal video understanding, tasks such as classification \cite{4}, segmentation \cite{5} and tracking \cite{6} also accomplished great achievements. Nevertheless, video and audio are separated when processing in the previous mentioned methods despite the fact of the two signals always coexist in real-world applications. 

There are plenty of visual and auditory signals from the world we live in. Our system is able to recognize objects and signals, then segment image regions covered by the objects, and ultimately isolate sounds produced by these objects in a joint fashion. Although lots of works have been put into the investigation of the correlation between synchronized visual and auditory signals, the applications are only focused on several coarse-grained tasks, such as event detection \cite{7}, sound separation \cite{8} and object localization \cite{9}. In many cases, the functioning aspects of the audio are overlooked. It is important to notice that the auditory signals contain semantic and positional information of the vocal objects and context of the background environment. These features have great potentials to be leveraged and utilized in dense prediction tasks over video domain. 

In this project, we successfully showed the great performance enhanced by the crossVis. On top of it, we tried to explore the joint learning scheme based on video and audio data. In particular, we investigated the correlation between video and audio in the context of video instance segmentation task. Initially, we sampled the audio spectrogram data according to the video frame rate. Afterwards, we managed to combine and assemble both of the signals and send each fused data to separate sets of FCN. After processed by the segmentation model, We measured the improvement in terms of the results of the segmentation.

The CrossVis \cite{18} uses the instance features in the current frame to localize the same instance in other frames by utilizing the FCOS(Fully Convolutional One State object detection). 

\section{Related Work}

\subsection{Video Instance Segmentation}
Recent internet world is engaged with massive amount of video data. Manually extracting semantic information from this enormous amount of internet video is highly unfeasible, seeking the need for automated methods to annotate useful information from the video data \cite{10}. Hence, one of the essential steps for video processing and retrieval is video segmentation. Video instance segmentation \cite{18,fu2020compfeat,li2021spatial,cao2020sipmask,li2022hybrid,li2022video} is an extended task from image segmentation \cite{he2017mask} with tracking instance identities. Video object segmentation \cite{yang2018efficient,jain2017fusionseg,huang2021scribble,zhang2018spftn,tokmakov2017learning} aims at partitioning every frame in a video into meaningful objects by grouping the pixels along spatio-temporal direction that exhibit coherency in appearance and motion \cite{11}. Video object segmentation task is highly challenging due to the following reasons: (i) unknown number of objects in a video (ii) varying background in a video and (iii) occurrence of multiple objects in a video \cite{12}. Existing approaches in video segmentation  can be broadly classified into two categories: interactive method and unsupervised method. With Interaction objects segmentation method, human intervention exists in initialization process while unsupervised approaches can perform object segmentation automatically. In semi-supervised approaches \cite{13}, user intervention is required for annotating initial frames and these annotations are transferred to the entire frames in the video. Recently, multimodal vision becomes more and more popular which uses text, audio or sensor data \cite{huang2021forgery,luo2021toward,li2020activitygan,huang2021towards} to facilitate the visual inputs. Some latest works focus on investigating multimodal video object segmentation which leverages text \cite{wu2022language}, audio \cite{li2022panoramic,li2022online} or wireless signals \cite{zhao2022self}.

\subsection{Audio-Visual Learning}
To learn the relationship between input sound and image, and better predict their representation, Zishun Feng and his team proposed an approach to explore the AVC performance not only on musical instrument data but also on videos in-the-wild. According to the paper \emph{Self-Supervised Audio-Visual Representation Learning for in-the-wild Videos.}\cite{15}
The authors used samples from video frames and 1-second audio segments from a given data-set of videos. The model was set up as an input pair of a single video frame and an audio segment, two ResNet-18 networks \cite{16} each to extract visual and audio features respectively. The activation was the distance between visual and audio features with a sigmoid function, finally to predict the correspondence of the input video frame and audio segment. The model behaved in a self-trained manner.
The authors trained their model using VGGSound dataset\cite{17}. The training parameters were 400k training steps with a batch size of 256. Over 100 million sampled audio-image pairs were used in total for training. The optimizer used was Adam Optimizer with a learning rate of 0.001.
The results showed that their method can be applied to obtain good retrieval results for video in various scenarios and achieved competitive results compared to other methods.
\section{Method}
\subsection{Model Description}
\begin{figure}[htbp]
\centering
\includegraphics[width=0.9\textwidth]{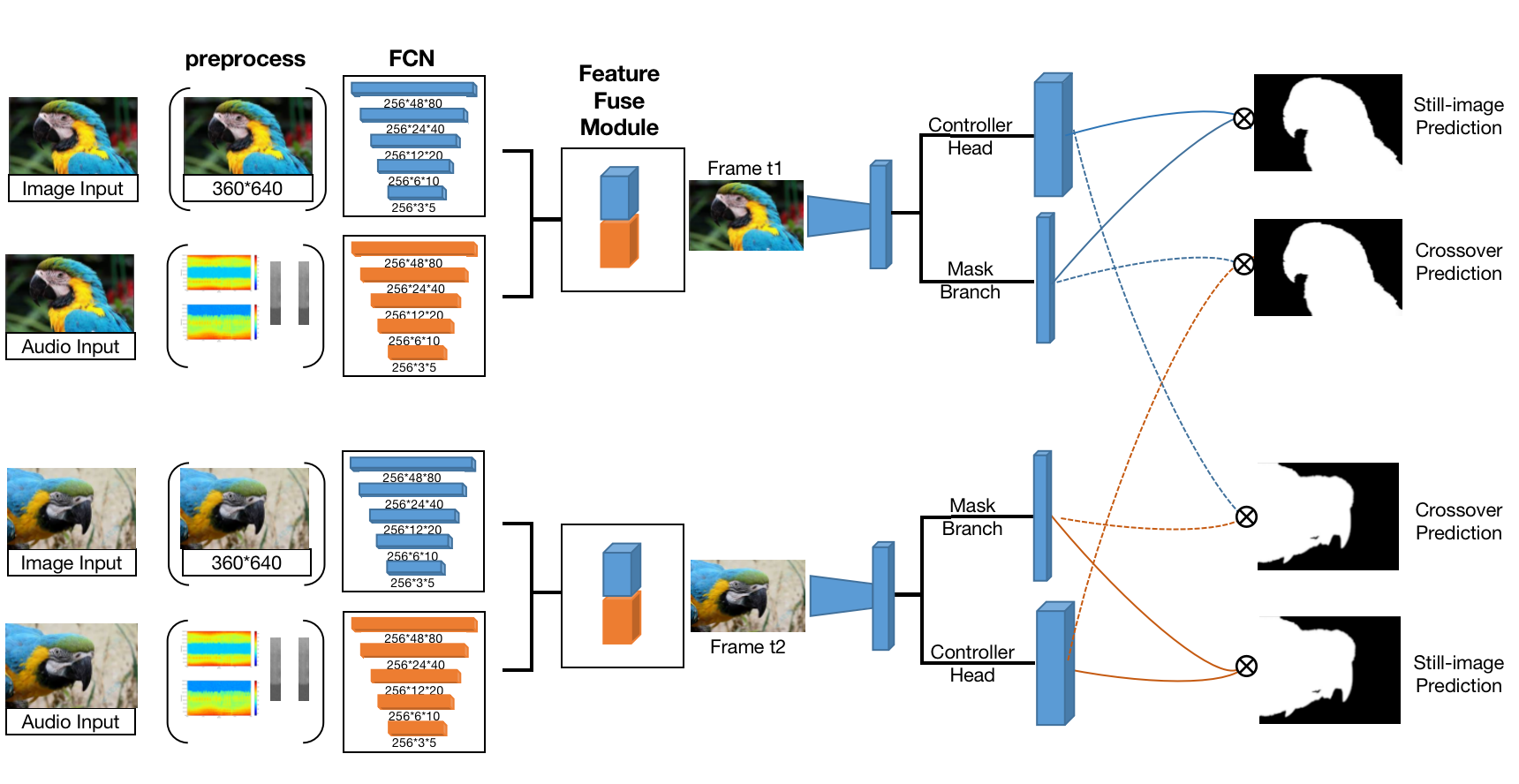}
\caption{\textbf{Audio Fusion CrossVis.} Our model which is based on CrossVis integrated audio features into the visual feature. The main trick we add on the Crossvis is that we used two baselines to extract both features from audio and visual signals. After extracted from signals, we concatenated these two feature maps and convolve them into one feature and then also do Crossover learning suggested as CrossVis paper.}
\label{fig:ourmodel}
\end{figure}

We integrated audio features with visual features in our model Figure~\ref{fig:ourmodel} with the methods used in Crossover Learning for Fast Online Video Instance Segmentation\cite{18} which is a Crossover Learning-based video instance segmentation framework. It achieves state-of-art performance among all published methods. We have built our method upon Crossover learning model and then we try to improve by adding audio cues in the following stage.
The proposed model is call CrossVis, and it consists of two key components: the crossover learning scheme for more accurate video-based instance representation learning and global balanced instance embedding branch. The authors originally leveraged the rich information across different video frames. Firstly, for still-image instance segmentation, we use dynamic conditional convolutions to generate the instance mask $\mathbf{M}_{x,y}$ by convolving a feature map $\tilde{\mathbf{F}}_{x,y}$from mask branch and a set of instance-specific dynamic filters $\mathbf{\theta}_{x,y}$

$$
\tilde{\mathbf{F}}_{x,y} = Concate(\mathbf{F}_{mask}; \mathbf{O}_{x, y})
$$
$$
\mathbf{M}_{x,y} = MaskHead(\tilde{\mathbf{F}}_{x,y}; \mathbf{\theta}_{x, y})
$$
Secondly, in terms of the VIS task, given a sampled frame-pair from one video, the same instance may appear in different locations. Thus, the authors utilize the appearance information $\theta_{x,y}$ from one sampled frame $t$ to incorporate the location information $O_{x,y}(t, \delta)$ of same instance in another sampled frame $t+\delta$.
Within each frame, at time $t$, the instance mask of $I_i(t)$ located at $(x,y)$ can be represented as:
$$\mathbf{M}_{x,y}(t) = MaskHead(\tilde{\mathbf{F}}_{x,y}(t);\mathbf{\theta}_{x,y}(t))$$At time $t + \delta$, the instance move from location $(x,y)$ to location $(x', y')$. So the instance mask of $I_i(t+\delta)$ can be represented as:
$$\mathbf{M}_{x',y'}(t+\delta) = MaskHead(\tilde{\mathbf{F}}_{x',y'}(t+\delta);\mathbf{\theta}_{x',y'}(t+\delta))$$Besides that, the model's trick crossover learning scheme establishes a connection between the dynamic filter from one frame and the mask feature map from another map. We expect the dynamic filter $\mathbf{\theta}_{x,y}(t)$can produce the mask of frame in another time $t$ by convolving its mask feature map $\tilde{\mathbf{F}}_{x',y'}(t+\delta)$:
$$\mathbf{M}^*_{x',y'}(t+\delta) = MaskHead(\tilde{\mathbf{F}}_{x',y'}(t+\delta);\mathbf{\theta}_{x',y'}(t))$$The same as mask in time t:$$\mathbf{M}^*_{x,y}(t) = MaskHead(\tilde{\mathbf{F}}_{x,y}(t);\mathbf{\theta}_{x',y'}(t+\delta))$$where $\mathbf{M}^*$ with a super script "*" denotes the instance mask produced by cross over learning. Following all four equations above, during training, we are all optimized by the dice loss:
$$
L_{dice}(\mathbf{M,M*}) = 1 -\frac{2\sum^{HW}_i\mathbf{M}_i \mathbf{M}_i^*}{\sum^{HW}_i(\mathbf{M}_i)^2 + \sum^{HW}_i(\mathbf{M}^*_i)^2}
$$



 
\section{Experiments}
\subsection{Datasets}
\begin{figure}[htbp]
\centering
\includegraphics[width=\linewidth, height=0.1\linewidth]{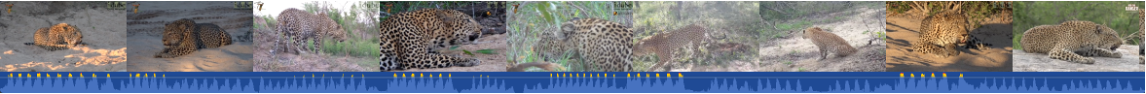}
\includegraphics[width=\linewidth, height=0.1\linewidth]{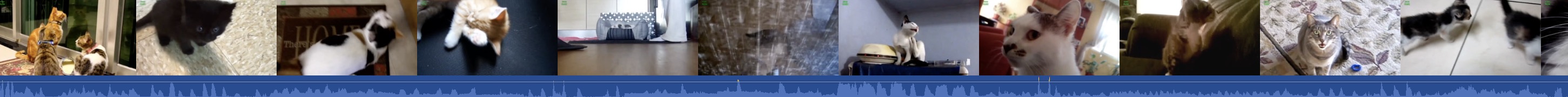}
\caption{\textbf{Dataset.} The first row and second row are leopard and cat categories respectively.}
\label{fig:dataset}
\end{figure}

We collected a dataset containing both videos as well as corresponding audio recordings. In particular, the dataset consists of two parts - Youtube-VIS-2019 (manually labeled), Self-collected dataset (auto-labeled). Youtube-VIS-2019 is a dataset for video instance segmentation based on initial Youtube-VOS dataset, with category label set including 40 common objects such person, animals and vehicles, and 4883 unique video instances and 131k high-quality manual annotations. Youtube-VIS-2019 have 2883 high-resolution YouTube videos which have been split into 2338 training videos, 302 validation videos and 343 test videos. Figure~\ref{fig:dataset} shows an example of our dataset.
\subsection{Data Preprocessing}
In order to get a better features from both visual and audio signals and experiment results, we did data preprocess for both visual and audio signals before feeding them into our model.

\subsubsection{Video}
First we extract picture frame from video at a certain frame rate. Then we resize all the pictures to the same size. Finally, we use ResizeShortestEdge function in object detection frame detectron2 to do data augmentation.

\subsubsection{Audio}
\begin{figure}[htbp]
\centering
\includegraphics[width=\linewidth]{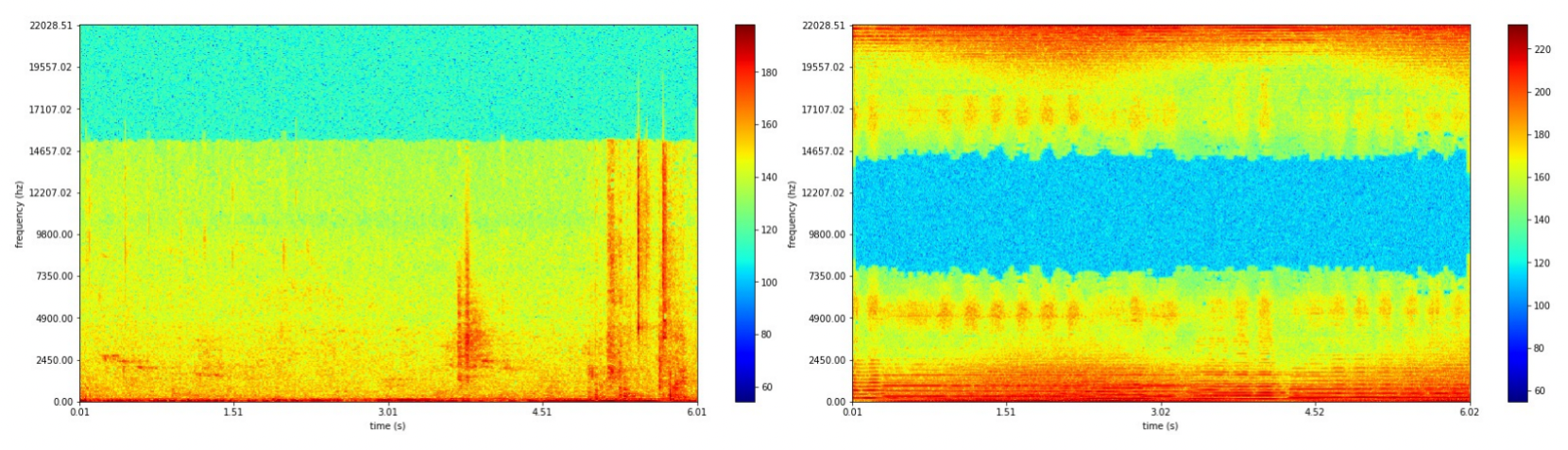}
\caption{Double/Single Audio Spectrogram.}
\label{fig:Double/Single Audio Spectrogram example}
\end{figure}

\begin{figure}[htbp]
\centering
\includegraphics[width=\linewidth]{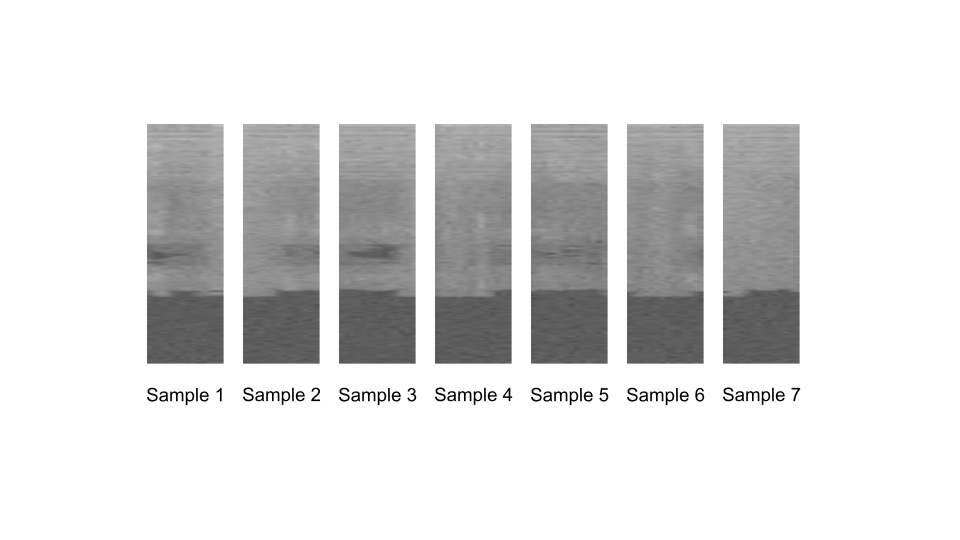}
\caption{Sampled Audio Spectrogram.}
\label{fig:Sampled Audio Apectrogram}
\end{figure}

We first convert the two channel audio into single channel audio in order to simplify future calculation. A spectrum Figure~\ref{fig:Sampled Audio Apectrogram} is a visual representation of the frequency spectrum of a signal as it changes over time. We preprocess the single channel audios into spectrum and remove those audios which don't have corresponding videos. Then according to the frame rate of the video, we evenly divide the spectrum along the X-axis which indicates time. The following pictures are double/single channel audio spectrogram and sampled audio Spectrogram. Figure~\ref{fig:Double/Single Audio Spectrogram example}

\subsection{Baseline Selection}
The CondInst instance segmentation model is pretrained on COCO train2017 with 1 * schedule following Detectron2 and AdelaiDet. And we initialize CrossVIS with it. We then train the CrossVIS on Youtube-VIS-2019 dataset with 1 * schedule, which refers to 12 epoch. The start learning rate is set to 0.005 according to SipMask-VIS. And at epoch of 9 and 11, learning rate is reduced by a factor of 10 respectively. For single-scale training, the frame is 360*640. And for multi-scale training, the frame is the same as the setting in SipMask-VIS. The frame is 360 * 640 during inference. We evaluate the CrossVIS on YouTube-VIS-2019. Unless specified, AP is defined as the area under the precision-recall (PR) curve. AP is averaged over multiple intersection-over-union (IoU) thresholds. And AR in this report refers to the average recall which is defined as the maximum recall given some fixed number of segmented instances per video. Following previous works, we report our results on the validation set to evaluate the effectiveness of the proposed method.
 
\subsection{Main Results}
\subsubsection{ Baseline model implementation}
\begin{figure}[htbp]
\centering
\includegraphics[width=0.9\textwidth]{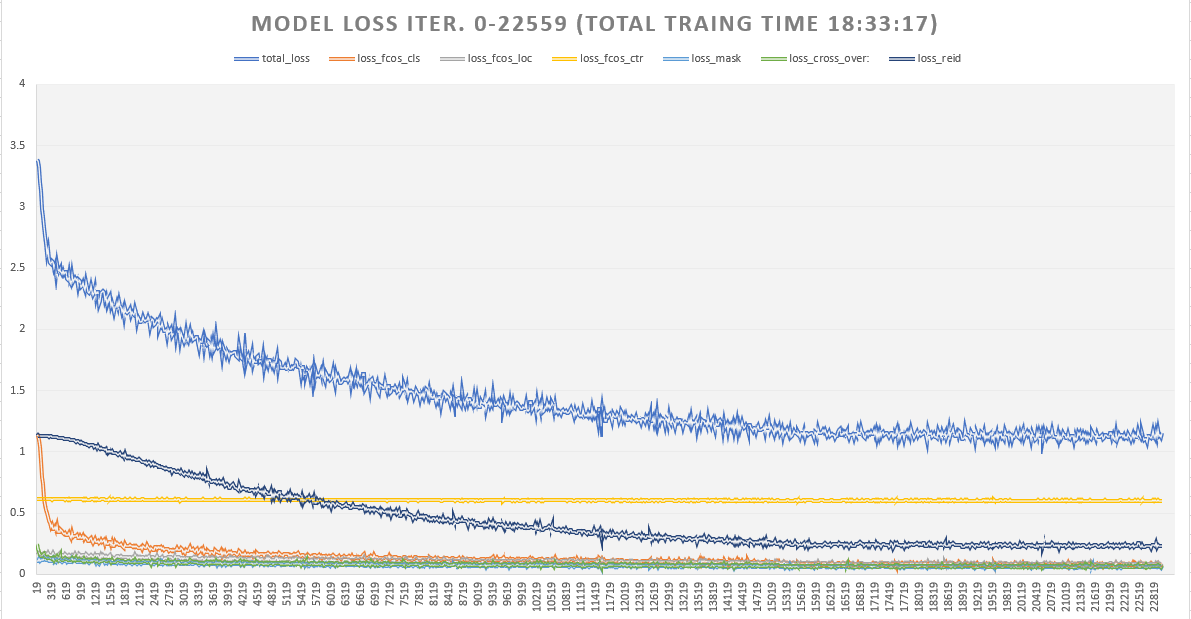}
\caption{\textbf{Training Loss.}.}
\label{fig:loss}
\end{figure}

At first, we implemented the CrossVis Model to get the segmentation result. After training 22559 iterations. as we can see from the Figure ~\ref{fig:loss}, the total loss has dropped to 1.132 with 0.08897 in focos classification loss, 0.008727 in focos localization loss loss,0.06239 in mask loss, 0.06859 in corss over learning loss, 0.2318 in embedding loss.
\\

\begin{figure}[htbp]
\centering
\includegraphics[width=0.9\linewidth]{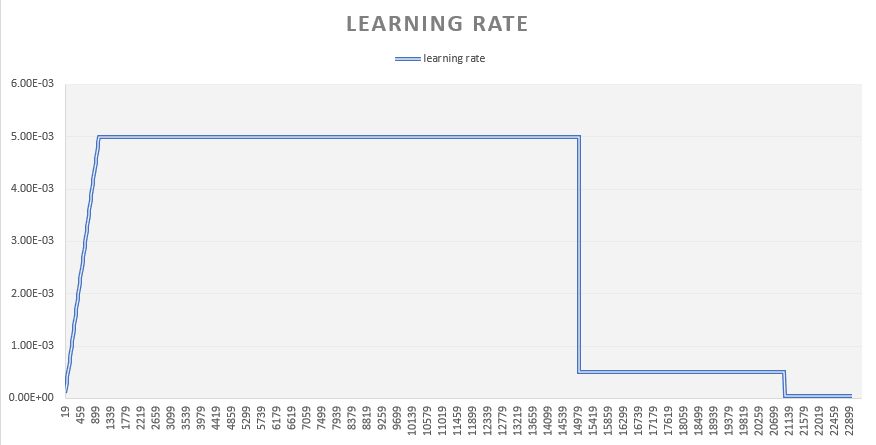}
\caption{\textbf{Learning Rate.}.}
\label{fig:learning rate}
\end{figure}

After implementation, we tried to tune the best parameters of the model. For example, We use WarmupMultiStepLR as the learning rate scheduler to train our model, which means it uses a small learning rate at first epochs. Through the stabilization of the model, we increase our learning rate to train. Our learning rate through one epoch shows as Figure ~\ref{fig:learning rate}

\subsubsection{Ablation Study}

\begin{figure}[htbp]
\centering
\includegraphics[width=\linewidth]{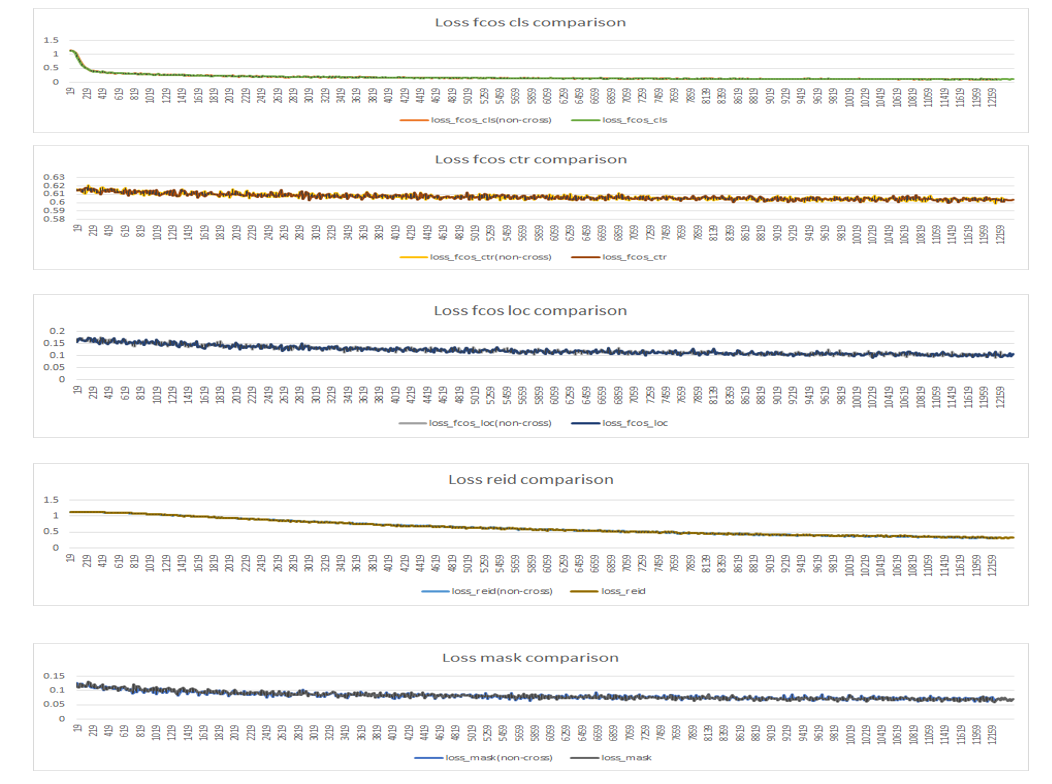}
\caption{\textbf{"CrossOver" Removal Experiment}.}
\label{fig: remvoal}
\end{figure}

From the paper of CrossVis, we learnt that the researchers try to use a novel crossover learning scheme that uses the instance features in the current frame to pixel-wisely localize the same instance in other frames.We'd like verify whether the "Cross" structure can really cause an improvement of Image Instance Segmentation Models.Therefore, we devised an experiment of  removing the "Cross" structure. The results of experiments is shown as Figure ~\ref{fig: remvoal}.)
\subsubsection{Audio Fusion CrossVis Performance}
\begin{figure}[htbp]
\centering
\includegraphics[width=\linewidth]{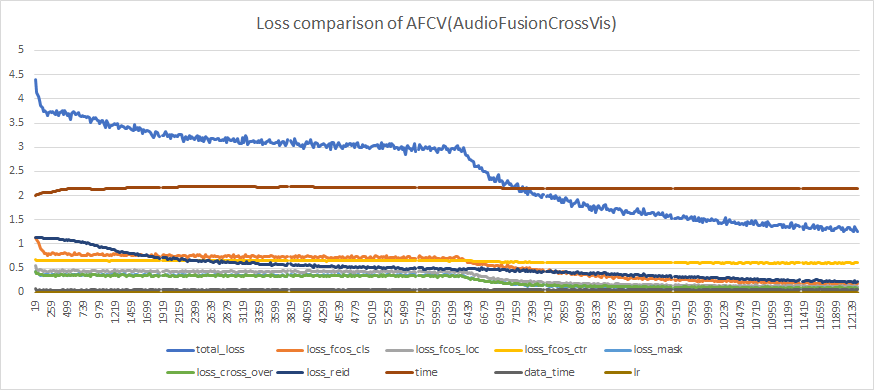}
\caption{\textbf{AFCV loss}.}
\label{fig: AFCV Loss}
\end{figure}

\begin{figure}[htbp]
\centering
\includegraphics[width=\linewidth]{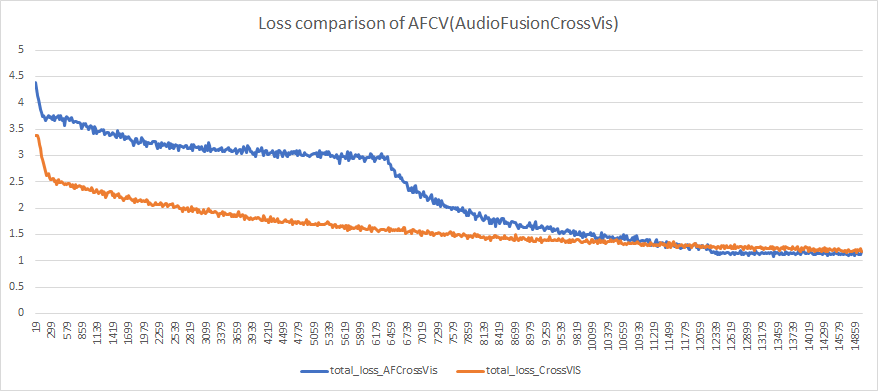}
\caption{\textbf{AFCV Convergence Performance}.}
\label{fig: AFCV performance}
\end{figure}

By introducing the features of audio, we formed our model Audio Fusion CrossVis. After training,  We got the loss as Figure ~\ref{fig: AFCV Loss} shows, we can see that total loss along with four individual losses decreases as we expect. It shows the convergence and proves the correctness of our model. 
Then we tried to compare the convergence rate of AFCV model and traditional CrossVis. The performance is shown as below Figure ~\ref{fig: AFCV performance}. Initially, the loss is higher than the baseline's due to new feature: audio's introduction. Over time, the loss of our model decreases then and after about 12000 iterations, the loss becomes stable and slightly smaller than the baseline's.

After verifying the correctness of the model and its' great performance on convergence, we tried to use evaluation metrics to compare its performance with other model like MaskTrack R-CNN, CondInst-VIS, CrossVis by introducing two datasets (COCO train 2017 and VIS datasets), What's more, We utilized Average Precision(AP) as our metrics to evaluate the performance. As we can see from the Table ~\ref{sample-table}, compared with MaskTack R-CNN, our model falls behind in terms of instance segmentation because of a rather shallow depth of mode. However, for the VIS tasks, our models shows a promising performance with a higher Average Precision over both the baseline model and MaskTrack R-CNN model. Therefore, we can conclude the effectiveness by integrating the audio's feature into our model.

\begin{table}
  \caption{Evaluation Metrics}
  \label{sample-table}
  \centering
  \begin{tabular}{llll}
    \toprule
    Method     & Backbone     & AP\textsuperscript{VIS} &AP\textsuperscript{COCO MASK} \\
    \midrule
    MaskTrack RCNN\cite{18} & ResNet-50  & 30.3 &34.7    \\
    MaskTrack RCNN\cite{18}     & ResNet-101  & 31.9 &35.9     \\
    CondInst-Vis     & ResNet-50  & 32.1 &35.7  \\
    CrossVis        & ResNet-50  & 34.8 &35.7 \\
    AFCV     & ResNet-50  & \textbf{34.9} &\textbf{35.7}\\
    \bottomrule
  \end{tabular}
\end{table}

\section{Conclusion}
We have successfully replicated the CrossVis framework solution with online video instance segmentation methods. And we have done the ablation experiment to test the effectiveness of crossover learning over CrossVIS framework. However, our experiment showed that there is limited improvements of performance of the cross learning over the baseline model. Then we fused the inputs from the audio aspect into the visual data, given the fact that the audio available from the sources of video are typically underutilized. By utilizing combined decoder to fuse both video and audio features, our model shows a promising improvements compared to the our baseline model. In the future, we plan to expand the size of our training set by annotating more objects in the Sound-20K dataset and self-collected dataset. Then we will evaluate alternative fusion methods, and perform sensitivity analysis on scaling factors and aspect ratios. And we will implement visual attention mechanism under the audio-visual condition, improving the accuracy of localization of sound-generating regions.

\bibliographystyle{plain}
\bibliography{ref.bib}
\end{document}